\documentclass[letterpaper, 10 pt, conference]{ieeeconf}  

\IEEEoverridecommandlockouts                              

\usepackage{microtype}      
\usepackage{xcolor}         
\usepackage{graphicx}
\usepackage{makecell}
\usepackage{booktabs}

\usepackage{amsmath}
\usepackage{multirow}
\newcommand{\eat}[1]{}
\newcommand{\warn}[1]{{\color{black}{#1}}}
\newcommand{\revise}[1]{{\color{black}{#1}}}

\title{\LARGE \bf
GANet: Goal Area Network for Motion Forecasting
}
\author{Mingkun Wang$^{1}$, Xinge Zhu$^{2}$, Changqian Yu$^{3}$, Wei Li$^{4}$, Yuexin Ma$^{5}$,\\ Ruochun Jin$^{6}$, Xiaoguang Ren$^{7}$, Dongchun Ren$^{3}$, Mingxu Wang$^{8}$ and Wenjing Yang$^{6*}$ 
\thanks{This work was supported by funding from the National Natural Science Foundation of China (91948303-1) and the National Key R\&D Program of China (2021ZD0140301).}
\thanks{*Corresponding author.}
\thanks{$^{1}$Peking University, wangmingkun95@qq.com}
\thanks{$^{2}$The Chinese University of Hong Kong, zhuxinge123@gmail.com}
\thanks{$^{3}$Meituan, yuchangqian@meituan.com, rendongchun@meituan.com}
\thanks{$^{4}$Inceptio, liweimcc@gmail.com}
\thanks{$^{5}$ShanghaiTech University, mayuexin@shanghaitech.edu.cn}
\thanks{$^{6}$National University of Defense Technology, wenjing.yang@nudt.edu.cn, jinrc@nudt.edu.cn}
\thanks{$^{7}$Academy of Military Sciences, rxg\_nudt@126.com}
\thanks{$^{8}$Fudan University, wang\_mingxu@126.com}
}

\begin{document}

\maketitle
\thispagestyle{empty}
\pagestyle{empty}

\begin{abstract}

Predicting the future motion of road participants is crucial for autonomous driving but is extremely challenging due to staggering motion uncertainty.
Recently, most motion forecasting methods resort to the goal-based strategy, i.e., predicting endpoints of motion trajectories as conditions to regress the entire trajectories, so that the search space of solution can be reduced.
However, accurate goal coordinates are hard to predict and evaluate.
In addition, the point representation of the destination limits the utilization of a rich road context, leading to inaccurate prediction results in many cases. Goal area, i.e., the possible destination area, rather than goal coordinate, could provide a more soft constraint for searching potential trajectories by involving more tolerance and guidance.  
In view of this, we propose a new goal area-based framework, named Goal Area Network (GANet), for motion forecasting, which models goal areas as preconditions for trajectory prediction, performing more robustly and accurately.
Specifically, we propose a GoICrop (Goal Area of Interest) operator to effectively aggregate semantic lane features in goal areas and model actors' future interactions as feedback, which benefits a lot for future trajectory estimations.
GANet ranks the \textbf{1st} on the leaderboard of Argoverse Challenge among all public literature (till the paper submission).
Code will be available at https://github.com/kingwmk/GANet.
\end{abstract}

\section{Introduction}

As one of the most critical subtasks in autonomous driving, motion forecasting targets \revise{to understand and predict} the future behaviors of other road participants (called actors).
It is essential for the self-driving car to make safe and reasonable decisions in the subsequent planning and control module.
The recent emergence of large-scale datasets with high-definition maps (HD maps) and sensor data~\cite{chang2019argoverse,wilson2021argoverse2, ettinger2021waymo} has boosted the research in motion forecasting. These HD maps provide rich geometric and semantic information, e.g., the map topology, that constrains the vehicle's motion. Meanwhile, actors also follow driving etiquette and 
interact with each other. Thus, how to effectively incorporate driving context to predict multiple plausible and accurate trajectories becomes the core challenge for motion forecasting.

Some works~\cite{casas2018intentnet,chai2019multipath} encode maps and motion trajectories into 2D images and apply convolutional neural networks (CNN) to process. 
Others~\cite{gao2020vectornet,zeng2021lanercnn} use vectorized and graph-structured data to represent maps.
For instance, LaneGCN~\cite{liang2020learning} applies a multi-stride graph neural network to encode maps.
However, since the traveling mode of an actor is highly diverse, the fixed size stride cannot effectively model distant relevant map features and thus limits the prediction performance (see Figure~\ref{vis}). 
While most works~\cite{mercat2020multi,casas2018intentnet,chai2019multipath,liang2020learning} focus on map encoding and motion history modeling, another family methods~\cite{zhao2020tnt,gu2021densetnt}, which is built on goal-based prediction, captures the actor's intentions in the future explicitly.
Specifically, these methods follow a three-stage scheme:
first, candidate goals are sampled from the lane centerlines;
second, a set of goals are selected by goal prediction;
third, trajectories are estimated conditioning on selected goals.
Although these methods have achieved competitive results, there remain two main drawbacks.
\revise{(1) These methods merely use a limited number of isolated goal coordinates as conditions, which contain limited information and hinder accurate motion forecasting.}
As goal coordinates of different distances to the road edge carry different information,
using a limited number of goal coordinates as conditions constrains the full utilization of a road context.
(2) The competitive performance of these methods heavily depends on the well-designed goal space, which may be violated in practice.
Well-designed goal space is required for sampling, refining, and scoring candidate goals,
due to the difficulty of predicting and evaluating accurate goal coordinates.
\revise{For example, vehicles' candidate goals are sampled from the lane centerlines while pedestrians' candidate goals are sampled from a virtual grid around themselves in TNT~\cite{zhao2020tnt}.}
However, these methods may fail once these hard-encoded candidate goals are violated in the real world. 

\eat{
First, the accurate goal coordinates are difficult to predict and evaluate.(How to connect? Because of the difficulty, they resort to sampling candidate goals(anchors), and refining and scoring them.) The competitive performance of these methods heavily depends on the well-designed goal space, which may be violated in practice.
\revise{For example, vehicles' candidate goals are sampled from the lane centerlines while pedestrians' candidate goals are sampled from a virtual grid around themselves in TNT~\cite{zhao2020tnt}.}
These hard-encoded candidate goals abide by driving rules, e.g., never depart far away from lanes or go beyond the road's edge, so the predicted trajectories are less likely to miss. 
However, these methods may fail, once these hard-encoded candidate goals are violated in the real world.
}

Compared with accurate goal coordinates, a potential goal area with a relatively richer context of the road is able to provide more tolerance and better guidance for accurate trajectory prediction through a soft constraint. 
Also, \warn{as driving history of actors is critical for goal area estimation}, we \eat{take advantage}\warn{make full use} of this \eat{cue}\warn{clue} for accurate localization of goal areas.
For example, a fast-moving vehicle's goal area may be far away, while the goal area of a stationary vehicle should be limited around itself.

\begin{figure*}
  \includegraphics[width=0.8\textwidth]{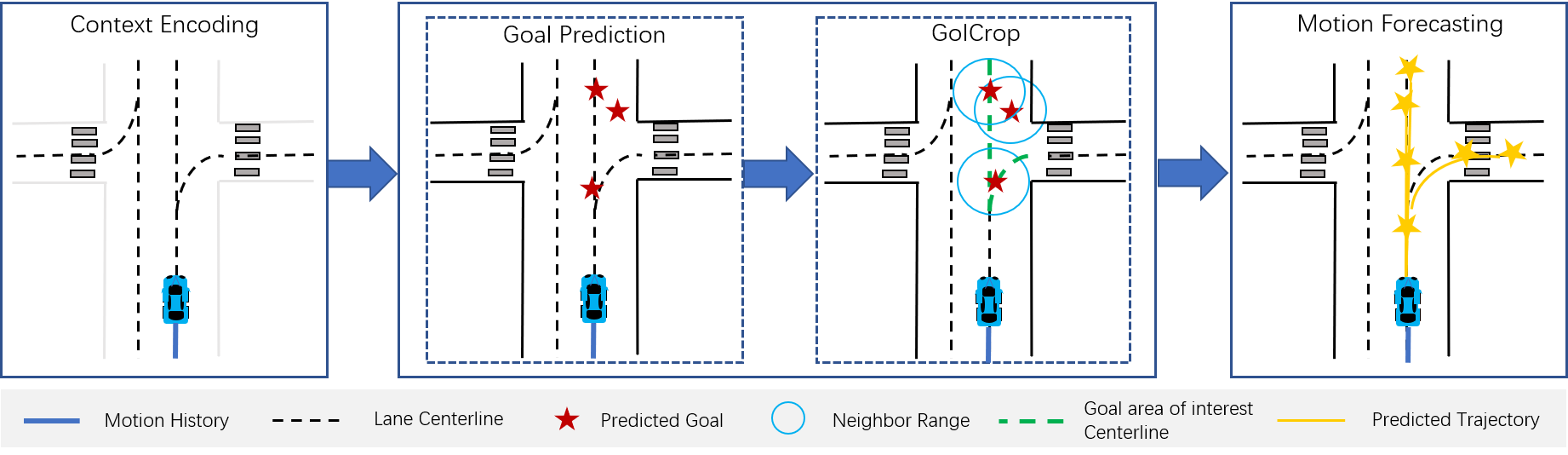}
  \centering 
  \caption{Illustration of GANet framework, which consists of three stages: (a) Context encoding encodes motion history and scene context; (b) Goal prediction predicts possible goals. GoICrop retrieves and aggregates goal area map features and models the actors' interactions in the future; (c) Motion forecasting estimates multi-feasible trajectories and their corresponding confidence scores.}
  \label{ts1}
  \vspace{-4ex}
\end{figure*}

Motivated by these observations, we propose \underline{G}oal \underline{A}rea \underline{Net}work framework (GANet)
that predicts potential goal areas as conditions for motion forecasting.
As shown in Figure \ref{ts1}, there are three stages in GANet, which are trained in an end-to-end way, and we construct a series of GANet models following this framework.
\revise{They overcome the shortcomings of the aforementioned goal-based prediction methods.}
First, \revise{an} efficient encoding backbone is adopted to encode motion history and scene context. 
Then, we predict approximate goals and crop their \eat{around}\warn{surrounding} goal areas as more robust conditions.
\warn{Moreover,} we introduce a GoICrop operator to explicitly query and aggregate the rich semantic features of lanes in the goal areas. 
Finally, we make the formal motion forecasting conditioned on motion history, scene context, and the aggregated goal area features.
Extensive experiments on the large-scale Argoverse 1  and Argoverse 2 motion forecasting benchmark demonstrate the effectiveness and generality of our proposed framework, where GANet achieves state-of-the-art performance.

\section{Related Work}
\noindent\textbf{Interactions.}~~~Early motion prediction methods mainly focus on motion and interaction modeling. 
They attempt to explain actors' complex movements by exploring their potential "interactions."
Traditional methods such as Social Force~\cite{helbing1995social} use hand-crafted features and rules to model interactions and constraints. 
Later, deep learning methods bring significant progress to this task. 
Social LSTM~\cite{alahi2016social} and SR-LSTM~\cite{zhang2019sr} use variants of LSTM to implicitly model interactions. 
GNN-TP~\cite{wang2019unsupervised} introduces a GNN method for interaction inference and trajectory prediction. 
The approach of~\cite{mercat2020multi} applies multi-head attention to incorporate interaction. mmTransformer~\cite{liu2021multimodal} applies a transformer architecture to fuse actors' motion histories, maps, and interactions. 

\noindent\textbf{HD maps encoding.}~~~According to the HD maps' processing manner, methods can be divided into three categories.
Rasterization-based methods rasterize the elements of HD maps and actors' motion histories into an image. 
Then, they use a CNN network to extract features and perform coordinates prediction. 
IntentNet~\cite{casas2018intentnet} develops a multi-task model with a CNN-based detector to extract features from rasterized maps. 
MultiPath~\cite{chai2019multipath} uses the Scene CNN to extract mid-level features and encodes the states of actors and their interactions on a top-down scene representation. However, these 2D-CNN-based methods suffer from low efficiency in extracting features of graph-structured maps.
Graph-based methods~\cite{zeng2021lanercnn} construct graph-structured representations from HD maps, which preserve the connectivity of lanes. 
VectorNet~\cite{gao2020vectornet} encodes map elements and actor trajectories as polylines and then uses a global interactive graph to fuse map and actor features. 
LaneGCN~\cite{liang2020learning} constructs a map node graph and proposes a novel graph convolution. 
Point cloud-based methods use points to represent actors' trajectories and maps. 
TPCN~\cite{ye2021tpcn} takes each actor as an unordered point set and applies a point cloud learning model.

\noindent\textbf{Multimodality.}~~~The multi-modal prediction has become an indispensable part of motion forecasting, which deals with the uncertainty in motion forecasting.
Generative methods, such as variational auto-encoder~\cite{lee2017desire} and generative adversarial network~\cite{gupta2018social}, can be used to generate multi-modal predictions.
However, each prediction requires independent sampling and forward pass, which cannot guarantee the diversity of samples.
Other methods~\cite{phan2020covernet,chai2019multipath} add some prior knowledge, such as pre-defined or model-based anchor trajectories. 
mmTransformer~\cite{liu2021multimodal} designs a region-based training strategy, which ensures that each proposal captures a specific pattern.
Recently, goal-based forecasting methods~\cite{zhang2020map} have proven effective. 
TNT~\cite{zhao2020tnt} first samples dense goal candidates along the lane and generates trajectories conditioned on high-scored goals. 
LaneRCNN~\cite{zeng2021lanercnn} regards each lane segment as an anchor. 
DenseTNT~\cite{gu2021densetnt} introduces a trajectory prediction model to output a set of trajectories from dense goal candidates. 
Heatmap-based methods~\cite{gilles2021gohome} focus on outputting a heatmap to represent the trajectories' future distribution. 
HOME~\cite{gilles2021home} method predicts a future probability distribution heatmap and designs a deterministic sampling algorithm for optimization. 

\textbf{Our method is different from previous works as follows.} (1) We give the definition of the goal area and propose a new goal area-based framework. We experimentally verify the effectiveness of modeling goal areas, predicting goal areas, and fusing crucial distant map features slighted by previous methods. These map features provide more robust information than the goal coordinates embedding.
(2) We employ a GoICrop operator to extract rich semantic map features in goal areas. It implicitly captures the interactions between maps and trajectories in goal areas and constrains the trajectories to follow driving rules and map topology in a data-driven manner. 
(3) Since our predicted goal is just a potential destination, we take it as a handle to model agents' interactions in the future, which is also crucial for collision avoidance.

\section{Method}
\begin{figure*}
  \includegraphics[width=0.9\textwidth]{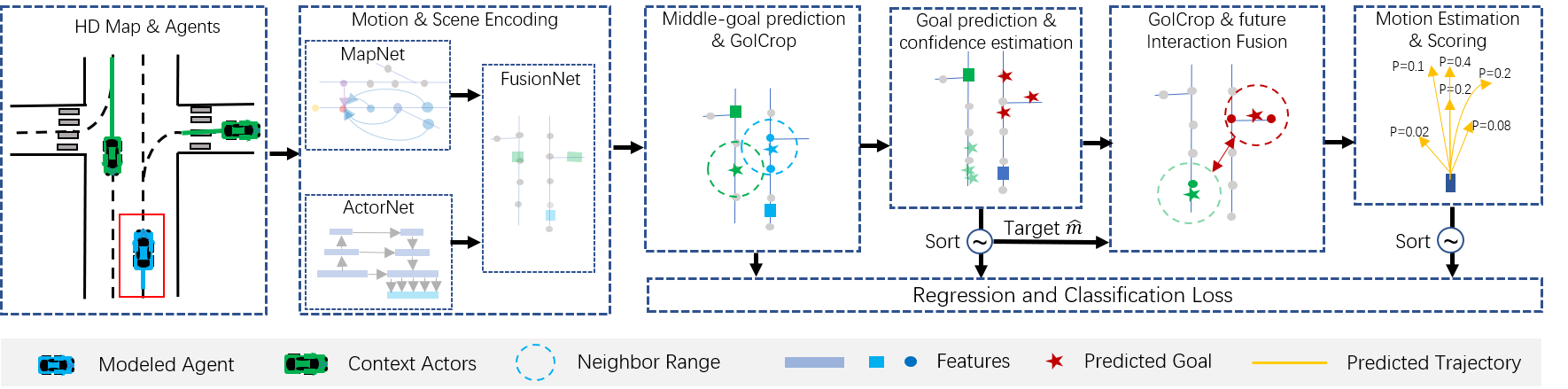}
  \centering 
  \caption{The GANet\_M\_3 model overview. (a) A feature extracting model encodes and fuses map and motion features. (b) The "one goal prediction" module predicts a goal area in the trajectory's middle position and aggregates its features. (c) The "three goals predictions" module predicts three goal areas, aggregates their features, and models the actors' future interactions. (d) The final prediction stage predicts $K$ trajectories and their confidence scores.}
  \label{ts2}
  \vspace{-4ex}
\end{figure*}

This section describes our formulation and GANet framework in a pipelined manner. An overview of the GANet architecture is shown in Figure~\ref{ts2}, and each module in this framework is pluggable. 

\textbf{Formulation.} Given a sequence of past observed states $a_{P}=[a_{-T^{'}+1},a_{-T^{'}+2},...,a_{0}]$ for an actor, we aim to predict its future states $a_{F}=[a_{1},a_{2},...,a_{T}]$ up to a fixed time step $T$. 
Running in a specific environment, each actor will interact with static HD maps $m$ and the other dynamic actors. Therefore, the probabilistic distribution we want to capture is $p(a_F|m, a_P, a^O_P)$, where $a^O_P$ denotes the other actors' observed states. The output of our model is $A_F = \{a_{F}^k\}_{k \in [0,K-1]}= \{(a_{1}^k,a_{2}^k,...,a_{T}^k)\}_{k \in [0,K-1]}$ for each actor, while motion forecasting tasks and subsequent decision modules usually expect us to output a set of trajectories. 

TNT-like methods' distribution can be approximated as
\begin{equation}
    \sum_{\tau \in T(m, a_P, a^O_P)}{p(\tau|m, a_P, a^O_P)p(a_F|\tau, m, a_P, a^O_P)}
\end{equation}
where $T(m, a_P, a^O_P)$ is the space of candidate goals depending on the driving context.
However, the map space $m$ is large, and the goal space $T(m, a_P, a^O_P)$ requires careful design.

Some methods expect to accurately predict the actor's motion by extracting good features.
For example, LaneGCN~\cite{liang2020learning} tries to approximate $p(a_F|m, a_P, a^O_P)$ by modeling $p(a_F|M_{a_0}, a_P, a^O_P)$, 
where $M_{a_0}$ is a "local" map features that is related to the actor's state $a_0$ at final observed step $t=0$.
To extract $M_{a_0}$, they use $a_0$ as an anchor to retrieve its surrounding map elements and aggregate their features. 
We found that not only the "local" map information is important, but also the goal area maps information is of great importance for accurate trajectory prediction. 
So, we reconstructed the probability as:
\begin{equation}
\sum _{\tau}{p(\tau|M_{a_0}, a_P, a^O_P) p(M_{\tau}|m, \tau)p(a_F|M_{\tau},M_{a_0}, a_P, a^O_P)}
\end{equation}
We directly predict possible goals $\tau$ based on actors' motion histories and driving context. Therefore, GANet is genuinely end-to-end, adaptive, and efficient. 
Then, we apply the predicted goals as anchors to retrieve the map elements in goal areas explicitly and aggregate their map features as $M_{\tau}$.

\subsection{Motion history and scene context encoding}
As shown in Figure~\ref{ts2}, the first stage of motion forecasting is driving context encoding, which extracts actors' motion features and maps features.
We adopt LaneGCN's~\cite{liang2020learning} backbone to encode motion history and scene context for its outstanding performance. 
Specifically, we apply a 1D CNN with Feature Pyramid Network (FPN) to extract actors' motion features.
Following~\cite{liang2020learning}, we use a multi-scale LaneConv network to encode the vectorized map data, which is consisted of lane centerlines and their connectivity. 
We construct a lane node graph from the map data. 
Finally, A fusion network transfers and aggregates feature among actors and lane nodes.
After driving context encoding, we obtain a 2D feature matrix $X$ where each row $X_i$ indicates the feature of the $i$-th actor, and a 2D matrix $Y$ where each row $Y_i$ indicates the feature of the $i$-th lane node.
We can also use other methods to encode motion history and scene context. For example, we implement a VectorNet++ method in the ablation study section. 

\subsection{Goal prediction}
In stage two, we predict possible goals for the $i$-th actor based on $X_i$. 
We apply intermediate supervision and calculate the smooth L1 loss between the best-predicted goal and the ground-truth trajectory's endpoint to backpropagate, making the predicted goal close to the actual goal as much as possible. 
The goal prediction stage serves as a predictive test to locate goal areas, which is different from goal-based methods using the predicted goals as the final predicted trajectories' endpoint. 
In practice, a driver's driving intent is highly multi-modal. 
For example, he or she may stop, go ahead, turn left, or turn right when approaching an intersection. 
Therefore, we try to make a multiple-goals prediction. 
We construct a goal prediction header with two branches to predict $E$ possible goals $G_{n,end} =\{g_{n,end}^e\}_{e \in [0,E-1]}$ and their confidence scores $    C_{n,end} = \{c_{n,end}^e\}_{e \in [0,E-1]}$, 
where $g_{n,end}^e$ is the $e$-th predicted goal coordinates and $c_{n,end}^e$ is the $e$-th predicted goal confidence of the $n$-th actor.

We train this stage using the sum of classification loss and regression loss.
Given $E$ predicted goals, we find a positive goal $\hat{e}$ that has the minimum Euclidean distance with the ground truth trajectory's endpoint. 
For classification, we use the max-margin loss:
\begin{equation}
    L_{cls\_end}=\frac{1}{N(E-1)}\sum_{n=1}^N\sum_{e\neq \hat{e}}{max(0,c^e_{n,end}+\epsilon -c^{\hat{e}}_{n,end})}
\end{equation}
where $N$ is the total number of actors and $\epsilon =0.2$ is the margin. The margin loss expects each goal to capture a specific pattern and pushes the goal closest to the ground truth to have the highest score.
For regression, we only apply the smooth L1 loss to the positive goals:
\begin{equation}
    L_{reg\_end}=\frac{1}{N}\sum_{n=1}^N{reg(g_{n,end}^{\hat{e}}-a^{*}_{n,end})}
\end{equation}
where $a^{*}_{n,end}$ is the ground truth BEV coordinates of the $n$-th actor trajectory's endpoint, $reg(z) = \sum_id(z_i)$, $z_i$ is the $i$-th element of $z$, and $d(z_i)$ is a smooth L1 loss.

Additionally, we also try to add a "one goal prediction" module at each trajectory's middle position aggregating map features to assist the endpoint goal prediction and the whole trajectory prediction.
Similarly, we apply a residual MLP to regress a middle goal $g_{n,mid}$ for the $n$-th actor. 
The loss term for this module is given by:
\begin{equation}
    L_{reg\_mid} = \frac{1}{N}\sum_{n=1}^N {reg(g_{n,mid}-a^*_{n,mid})}
\end{equation}
where $a^*_{n,mid}$ is the ground truth BEV coordinates of the $n$-th actor trajectory's middle position.

The total loss at the goal prediction stage is:
\begin{equation}
    L_{1} = \alpha_1 L_{cls\_end} + \beta_1 L_{reg\_end} +\rho_1 L_{reg\_mid}
\end{equation}
where $\alpha_1 = 1$, $\beta_1 = 0.2$ and $\rho_1 = 0.1$.

\subsection{GoICrop}
We choose the predicted goal with the highest confidence among $E$ goals as an anchor. This anchor is the approximate destination with the highest possibility that the actor may reach based on its motion history and driving context.
Because the actors' motion is highly uncertain, we crop maps within 6 meters of the anchor as the goal area of interest, which relaxes the strict goal prediction requirement. The actual endpoint is more likely to appear in candidate areas compared with being hit by scattered endpoint predictions.
Moreover, the actor's behavior highly depends on its destination area's context, i.e., the maps and other actors. Although previous works have explored the interactions between actors, the interactions between actors and maps in goal areas and the interactions among actors in the future have received less attention. 
Thus, we retrieve the lane nodes in goal areas and apply a GoICrop module to aggregate these map node features as follows:
\begin{equation}
    x'_i = \phi_1(x_iW_0+\sum_j\phi_2(concat(x_iW_1,\Delta_{i,j},y_j)W_2))W_3
    \label{att}
\end{equation}
where $x_i$ is the feature of $i$-th actor and and $y_j$ is the feature of $j$-th lane node, $W_i$ is a weight matrix, $\phi_i$ is a layer normalization with ReLU function, and $\Delta_{i,j}=\phi(MLP(v_i-v_j))$, where $v_i$ denotes the anchor's coordinates of $i$-th actor and $v_j$ denotes the $j$-th lane node's coordinates. 
GoICrop serves as spatial distance-based attention and updates the goal area lane nodes' features back to the actors. We transpose $x_i$ with $W_1$ as a query embedding. The relative distance feature between the anchor of $i$-th actor and $j$-th lane node are extracted by $\Delta_{i,j}$. Then, we concatenate the query embedding, relative distance feature, and lane node feature. An $MLP$ is employed to transpose and encode these features. Finally, the goal area features are aggregated for $i$-th actor.

Previous motion forecasting methods usually focus on the interactions in the observation history. 
However, actors will interact with each other in the future to follow driving etiquette, such as avoiding collisions. 
Since we have performed predictive goal predictions and gotten possible goals for each actor, our framework can model the actors' future interactions.
Hence, we utilize the predicted anchor positions and apply a GoICrop module as equation~\ref{att} to implicitly model actors' interactions in the future. We consider the other actors whose future anchor's distance from the anchor of $i$-th actor is smaller than 100 meters. In this case, $y_j$ in equation~\ref{att} denotes the features of $j$-th actor, $v_i$ denotes the anchor's coordinates of $i$-th actor, and $v_j$ denotes the anchor's coordinates of $j$-th actor in $\Delta_{i,j}=\phi(MLP(v_i-v_j))$.

\begin{table*}\small
\caption{Results on Argoverse 1 (upper set) and Argoverse 2 (lower set) motion forecasting test dataset. The "-" denotes that this result was not reported in their paper.}
\label{test}
\centering
\begin{tabular}{cccccccc}
\toprule
Method     &\makecell[c]{b-minFDE\\(K=6)} &\makecell[c]{MR\\(K=6)} &\makecell[c]{minFDE\\(K=6)} &\makecell[c]{minADE\\(K=6)} &\makecell[c]{minFDE\\(K=1)} &\makecell[c]{minADE\\(K=1)} &\makecell[c]{MR\\(K=1)} \\
\midrule
LaneRCNN \cite{zeng2021lanercnn} &2.147  &0.123 &1.453  &0.904 &3.692 &1.685 &0.569     \\
TNT\cite{zhao2020tnt} &2.140 &0.166 &1.446 &0.910 &4.959 &2.174  &0.710      \\

DenseTNT (MR)\cite{gu2021densetnt}	&2.076	&0.103 &1.381 & 0.911	&3.696	&1.703 &0.599 \\

\textbf{\emph{LaneGCN~\cite{liang2020learning}}} &\emph{2.059}	&\emph{0.163} &\emph{1.364}	&\emph{0.868}	&\emph{3.779}	&\emph{1.706}  &\emph{0.591}  \\

mmTransformer\cite{liu2021multimodal}   &2.033	&0.154 &1.338 & 0.844 &4.003	&1.774 &0.618\\

GOHOME~\cite{gilles2021gohome}      &1.983	&0.105 &1.450 &0.943	&3.647 &1.689 &0.572 \\

HOME \cite{gilles2021home}  &-	&\textbf{0.102} &1.45	&0.94	&3.73	&1.73	&0.584  \\

DenseTNT (FDE)\cite{gu2021densetnt}   	&1.976	&0.126 &1.282	&0.882	&3.632	&1.679 &0.584\\

TPCN~\cite{ye2021tpcn}           &1.929	&0.133 &1.244	&0.815	&3.487	&\textbf{1.575} &0.560	  \\

        
\textbf{GANet(Ours)}    &\textbf{1.790}	&0.118 &\textbf{1.161}	&\textbf{0.806} &\textbf{3.455} &1.592 &\textbf{0.550}	 \\
\midrule

DirEC &3.29  &0.52 &2.83 &1.26 &	6.82 &	2.67 &0.73    \\

drivingfree&3.03 &	0.49 &2.58 &1.17 &6.26 & 2.47 &0.72     \\

LGU	&2.77 & 0.37 & 2.15 &1.05 &6.91 & 2.77 & 0.73 \\

Autowise.AI(GNA) &2.45	&0.29 &1.82	&0.91 &6.27	& 2.47  &	0.71\\

Timeformer~\cite{gilles2022thomas}   &2.16	&0.20 &1.51 & 0.88 &4.71	&1.95 &0.64\\

QCNet    &2.14	&0.24 &1.58 &0.76	&4.79 & 1.89 &0.63 \\

\textit{ OPPred w/o Ensemble}~\cite{zhang2022oppred} 	&2.03	&0.180 & 1.389		&0.733	&4.70	&1.84  &0.615\\

\textit{TENET w/o Ensemble}~\cite{wang2022tenet}&2.01&- &-	&-	&-	&-&-\\

Polkach(VILaneIter)  &2.00	&0.19 & 1.39	&\textbf{0.71}	&4.74	&1.82	&0.61 \\

\textbf{GANet(Ours)}    &\textbf{1.969}	& \textbf{0.171} &\textbf{1.352}	&0.728 &\textbf{4.475} &\textbf{1.775} &\textbf{0.597}	 \\
\bottomrule
\vspace{-7ex}
\end{tabular}
\end{table*}

\subsection{Motion estimation and scoring}
We take the updated actor features $X$ as input to predict $K$ final future trajectories and their confidence scores in stage three. 
Specifically, we construct a two-branch multi-modal prediction header similar to the goal prediction stage, with one regression branch estimating the trajectories and one classification branch scoring the trajectories.
For each actor, we regress $K$ sequences of BEV coordinates 
$    A_{n,F}=\{(a_{n,1}^k,a_{n,2}^k,...,a_{n,T}^k)\}_{k \in [0,K-1]}$,
where $a_{n,t}^k$ denotes the $n$-th actor's future coordinates of the $k$-th mode at $t$-th step. 
For the classification branch, we output $K$ confidence scores $    C_{n,cls} = \{c_n^k\}_{k \in [0,K-1]}
$ corresponding to $K$ modes.
We find a positive trajectory of mode $\hat{k}$, whose endpoint has the minimum Euclidean distance with the ground truth endpoint.

For classification, we use the margin loss  $L_{cls}$ similar to the goal prediction stage. 
For regression, we apply the smooth L1 loss on all predicted steps of the positive trajectories:
\begin{equation}
    L_{reg}=\frac{1}{NT}\sum_{n=1}^N\sum_{t=1}^T{reg(a_{n,t}^{\hat{k}}-a^{*}_{n,t})}
\end{equation}
where $a^{*}_{n,t}$ is the $n$-th actor's ground truth coordinates.

To emphasize the importance of the goal, we add a loss term stressing the penalty at the endpoint:
\begin{equation}
    L_{end}=\frac{1}{N}\sum_{n=1}^N{reg(a_{n,end}^{\hat{k}}-a^{*}_{n,end})}
\end{equation}
where $a^{*}_{n,end}$ is the $n$-th actor's ground truth endpoint coordinates and $a_{n,end}^{\hat{k}}$ is the $n$-th actor's predicted positive trajectory's endpoint.

The loss function for training at this stage is given by:
\begin{equation}
    L_{2} = \alpha_2 L_{cls} + \beta_2 L_{reg} + \rho_2 L_{end}
\end{equation}
where $\alpha_2=2$, $\beta_2=1$ and $\rho_2=1$.

\subsection{Training}
As all the modules are differentiable, we train our model with the loss function:
\begin{equation}
L = L_{1} + L_{2}
\end{equation}
The parameters are chosen to balance the training process.

\section{Experiments}
\subsection{Experimental settings}
\noindent\textbf{Dataset.}~~~Argoverse 1~\cite{chang2019argoverse} is a large-scale motion forecasting dataset, which consists of over 30K real-world driving sequences,
\revise{split into} train, validation, and test sequences without geographical overlap. 
\revise{Each training and validation sequence is 5 seconds long,
while each test sequence presents only 2 seconds to the model, and another 3 seconds are withheld for the leaderboard evaluation.} Each sequence includes one interesting tracked actor labeled as the "agent."
Given an initial 2-second observation, the task is to predict the agent's future coordinates in \revise{the} next 3 seconds.

Spanning 2,000+ km over six geographically diverse cities, 
Argoverse 2~\cite{wilson2021argoverse2} is a high-quality motion forecasting dataset whose scenario is paired with a local map. Each scenario is 11 seconds long. We observe five seconds and predict six seconds for the leaderboard evaluation.
Compared to Argoverse 1, the scenarios in Argoverse 2 are approximately twice longer and more diverse.

\noindent\textbf{Metrics.}~~~\revise{We follow the widely used evaluation metrics~\cite{zeng2021lanercnn,gu2021densetnt,ye2021tpcn}.} 
\revise{Specifically,} MR is the ratio of predictions where none of the predicted $K$ trajectories is within 2.0 meters of ground truth according to the endpoint's displacement error. 
Minimum Final Displacement Error (minFDE) is the L2 distance between the endpoint of the best-forecasted trajectory and the ground truth. 
Minimum Average Displacement Error (minADE) is the average L2 distance between the best-forecasted trajectory and the ground truth. 
Argoverse Motion Forecasting leaderboard is ranked by Brier minimum Final Displacement Error (brier-minFDE6), which adds a probability-related penalty to the endpoint's L2 distance error. 

\noindent\textbf{Implementation.}~~~ We train our model on 2 A100 GPUs using a batch size of 128 with the Adam optimizer for 42 epochs. The initial learning rate is 1 x 10-3, decaying to 1 x 10-4 at 32 epochs. 

\begin{figure*}[htb]
  \includegraphics[width=0.8\textwidth]{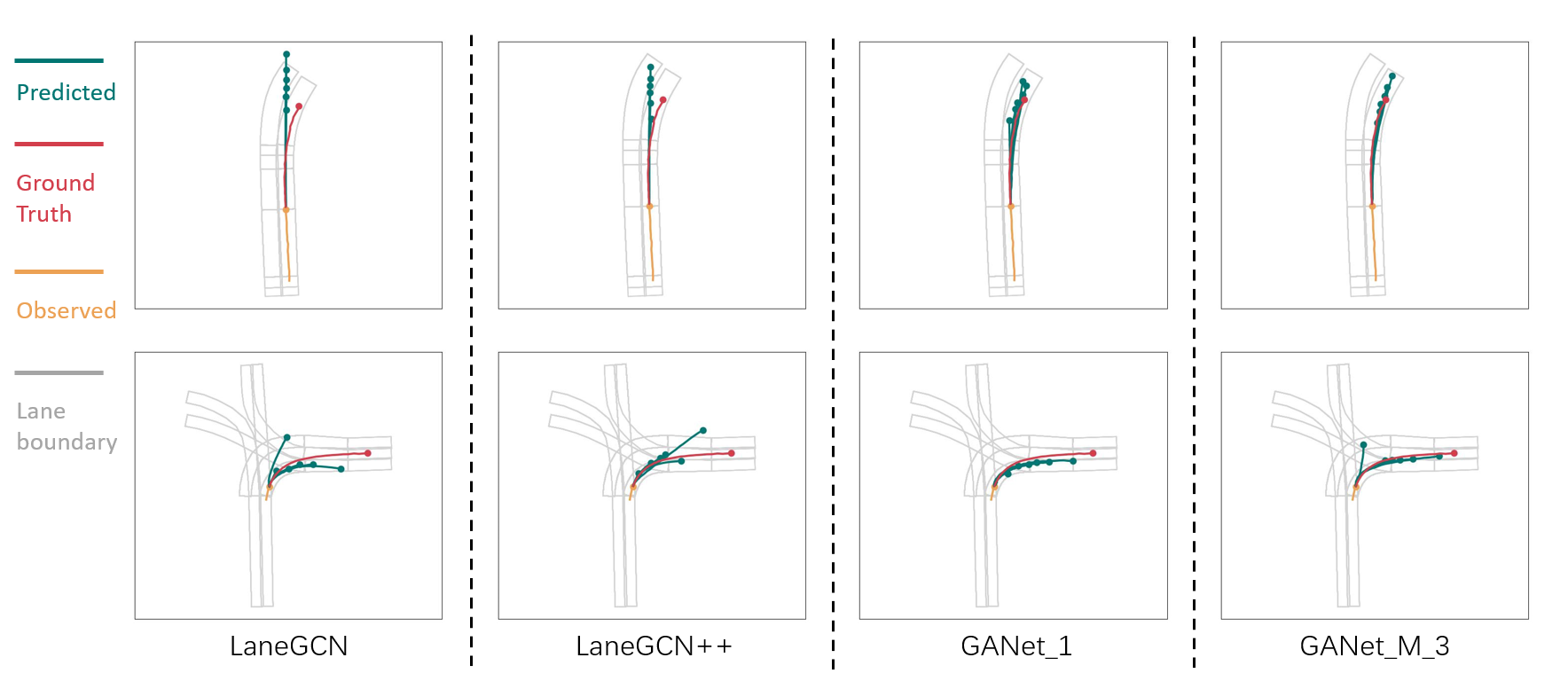}
  \centering 
  \caption{Qualitative results on the Argoverse 1 validation set. Lanes are shown in grey, the agent's past trajectory is in orange, the ground truth future trajectory is in red, and the predicted six trajectories are in green. The results of different methods are shown in different columns.}
  \label{vis}
  \vspace{-2ex}
\end{figure*}

\subsection{Comparison with State-of-the-art}
We compare our approach with \eat{current}state-of-the-art methods. 
As shown in Table~\ref{test}, our GANet outperforms \eat{the}existing goal-based approaches such as TNT~\cite{zhao2020tnt}, LaneRCNN~\cite{zeng2021lanercnn}, and DenseTNT~\cite{gu2021densetnt}.
\revise{Specifically, we make a detailed comparison with LaneGCN because we adopt their backbone to encode motion history and scene context.}
\revise{Public results on the official motion forecasting challenge leaderboard show that our GANet method significantly beats LaneGCN by decreases of 28\%, 15\%, 13\% and 9\% in MR6, minFDE6, brier-minFDE6, and minFDE1, respectively, which demonstrate the effectiveness of GANet.}
We also conduct experiments on Argoverse 2 Motion Forecasting Dataset~\cite{wilson2021argoverse2}, and GANet is the winner that achieves state-of-the-art performance in CVPR 2022 Argoverse Motion Forecasting Challenge, whose top ten entries are shown in Table~\ref{test}. Since many methods, such as TENET and OPPred, apply model ensemble to boost their performance, we report their results without an ensemble for a fair comparison.


\begin{table}\small
\caption{Ablation study results on the Argoverse 1 validation set.}
\label{ablation1}
\centering
\begin{tabular}{c|c|cccc}
\toprule
Method
&\makecell[c]{minFDE\\(K=6)}  &\makecell[c]{minADE\\(K=6)} &\makecell[c]{minFDE\\(K=1)}  &\makecell[c]{minADE\\(K=1)} \\
\midrule
LaneGCN 	            &1.080	&0.710	&3.010	&1.359    \\
LaneGCN++     	&1.076	&0.703	&2.819	&1.286    \\
GANet\_1 	        &0.961	&0.684	&2.743	&1.269    \\
GANet\_3      	&0.949	&0.679	&2.719	&1.264    \\
GANet\_M\_3   	&\textbf{0.934}&\textbf{0.673}&\textbf{2.707} &\textbf{1.259}\\
\midrule
GANet\_2 	    &0.971	&0.689	&2.756	&1.280  \\
GANet\_6	&0.966	&0.683	&2.784	&1.289  \\
GANet\_9	&0.967	&0.685	&2.759	&1.282  \\
\midrule
VectorNet~\cite{gao2020vectornet}   &-	&-	&3.67	&1.66   \\
VectorNet++  &1.156	&0.772	&3.256	&1.507   \\
GANet\_1 	& 1.076 & 0.744& \textbf{3.050} & \textbf{1.429}\\
GANet\_M\_3    & \textbf{1.042} & \textbf{0.732} & 3.100 & 1.449\\
\bottomrule
\end{tabular}
\vspace{-5ex}
\end{table}

\subsection{Ablation studies}
\noindent\textbf{Component study.}~~~\revise{We perform ablation studies on the validation set to investigate the effectiveness of each component.}
Taking the LaneGCN model as a baseline, we add other components progressively. 
First, to emphasize the motion's temporal modeling, we construct an enhanced version of LaneGCN called LaneGCN++. 
Specifically, we apply an LSTM network on FPN's output features and use two identical parallel networks to enhance the motion history encoding.
As shown in Table~\ref{ablation1}, LaneGCN++ improves the ADE1 and FDE1 metrics' performance. 
However, the enhanced bigger network shows little improvement in multi-modal prediction.

Second, to verify GANet's effectiveness, we adopt LaneGCN++'s backbone and add a "one goal prediction" module to construct the GANet\_1 model, which only predicts $M=1$ goals. 
Since we only predict one goal in this model, we omit the classification loss term $L_{cls\_end}$ and $L_{reg\_mid}$ in $L_{1}$. 
The performance of the GANet\_1 model outperforms LaneGCN++ dramatically, with more than 10\% improvement on minFDE6. 
In addition, considering the multimodality, we apply a "three goals predictions" module in our GANet\_3 model, which performs better. 
Moreover, we also try to add a "one goal prediction" module at the trajectory's middle position to aggregate the middle position's map information in GANet\_M\_3.
The performance has been further improved.
Our models improve all the metrics compared to the LaneGCN++.

\noindent\textbf{Number of goals.}~~~We also evaluate the effect of \revise{the goal number}. \revise{Table \ref{ablation1} shows the model performance under different numbers of goals, where the goal number only has marginal effects on the overall performance.}\eat{ Although the number of goals has an effect, it does not make a big difference.} 

\noindent\textbf{Backbone.}~~~To demonstrate the generality of GANet, we implement a VectorNet++ method \revise{as another backbone}, whose polylines idea is similar to VectorNet~\cite{gao2020vectornet}. 
We construct our GANet models adopting the VectorNet++ backbone. As shown in Table~\ref{ablation1}, the performance improves by 9.9\% and 5.2\% in minFDE6 and minADE6, respectively, which shows the generality of GANet when adopting different scene context encoding methods.


\subsection{Qualitative results}

We visualize the predicted results on the validation set.
\revise{For challenging sequences, almost all results of GANet models are more reasonable and smoother following map constraints than outputs of LaneGCN.}
We show the multi-modal prediction of two cases in Figure \ref{vis} and compare GANet with LaneGCN qualitatively. 
\revise{For illustration purpose,} we only draw the agent's trajectory for an intuitive check while other actors are omitted.
The first row shows a case where the direction of the lane has changed over a long distance. 
LaneGCN\eat{'s method} is unaware of this distant change and gives six straight predictions. 
GANet\_1 model captures this change and generates trajectories that follow the lane topology, while GANet\_M\_3 model generates smoother trajectories than GANet\_1\eat{ model}.
The second row \eat{shows}\revise{presents} a case where the agent performs a right turn at a complex intersection. 
Due to the \eat{lacking}\revise{lack} of motion history, maps are essential to produce reasonable trajectories.
LaneGCN produces divergent, non-traffic-rule compliant trajectories, while our \eat{model}\revise{method} produces reasonable trajectories following the lane topology. 


\section{Conclusion}
This paper proposes a Goal Area Network~(GANet), a new framework for motion forecasting.
GANet predicts potential goal areas as conditions for prediction. 
We design a GoICrop operator to extract and aggregate the rich semantic lane features in goal areas.
It implicitly models the interactions between trajectories and maps in the goal area and the interactions between actors in the future in a data-driven manner. 
Experiments on the Argoverse motion forecasting benchmark demonstrate GANet's effectiveness.

\addtolength{\textheight}{-4cm}   
\end{document}